\documentclass[a4paper,fleqn]{cas-dc}

\usepackage[numbers,sort&compress]{natbib}
\usepackage{graphicx}
\usepackage{threeparttable}
\usepackage{amsmath}
\usepackage{amssymb}
\usepackage{booktabs}
\usepackage{amsfonts}
\usepackage{multirow}
\usepackage{multicol}
\usepackage{tcolorbox}
\usepackage[scaled=.90]{helvet}
\usepackage{caption}
\DeclareCaptionFont{xipt}{\fontsize{9}{11}\sf\selectfont}
\newcommand{\mygreen}[1]{\textbf{\tiny \tcbox[size=fbox, colback=green!30, colframe=green!30!black]{#1}}}
\newcommand{\myred}[1]{\textbf{\tiny \tcbox[size=fbox, colback=red!40, colframe=red!40!black]{#1}}}
\newcommand{\mygray}[1]{\textbf{\tiny \tcbox[size=fbox, colback=black!10, colframe=black!10!black]{#1}}}

\newcommand{\teblebold}[1]{\footnotesize\textbf{#1}}

\ExplSyntaxOn
\cs_gset:Npn \__first_footerline:
  { \group_begin: \small \sffamily \__short_authors: \group_end: }
\ExplSyntaxOff

\def\tsc#1{\csdef{#1}{\textsc{\lowercase{#1}}\xspace}}
\tsc{WGM}
\tsc{QE}
\tsc{EP}
\tsc{PMS}
\tsc{BEC}
\tsc{DE}

\newlength\savewidth
\newcommand{\tablestyle}[2]{\setlength{\tabcolsep}{#1}\renewcommand{\arraystretch}{#2}\centering\footnotesize}
\begin{document}
\let\WriteBookmarks\relax
\def\floatpagepagefraction{1}
\def\textpagefraction{.001}

\shorttitle{Engineering Science and Technology, an International Journal}

\shortauthors{O.N Manzari et~al.}

\title [mode = title]{Robust Transformer with Locality Inductive Bias and Feature Normalization}


\author[1]{Omid Nejati Manzari}[type=editor,
                        auid=000,bioid=1,
                        ]
\cormark[1]
\ead{omid_nejaty@alumni.iust.ac.ir}

\credit{Conceptualization, Software, Writing- Original draft preparation, Validation, Resources}

\address[1]{School of Electrical Engineering, Iran University of Science and Technology, Tehran, Iran}
\address[2]{Lane Department of Computer Science and Electrical Engineering, West Virginia University, Morgantown, USA}

\author[2]{Hossein Kashiani}[
   ]
\ead{hk00014@mix.wvu.edu}

\credit{Methodology, Writing- Reviewing and Editing}

\author[1]
{Hojat Asgarian Dehkordi} 
\ead{h_asgariandehkordi@elec.iust.ac.ir}

\credit{Modification for the final layout}

\author[1]{Shahriar Baradaran Shokouhi} 
\ead{bshokouhi@iust.ac.ir}

\credit{Supervision, Review \& Editing}

\cortext[cor1]{Corresponding author.}

\begin{abstract}
Vision transformers have been demonstrated to yield state-of-the-art results on a variety of computer vision tasks using attention-based networks. However, research works in transformers mostly do not investigate robustness/accuracy trade-off, and they still struggle to handle adversarial perturbations. In this paper, we explore the robustness of vision transformers against adversarial perturbations and try to enhance their robustness/accuracy trade-off in white box attack settings. To this end, we propose Locality iN Locality (LNL) transformer model. We prove that the locality introduction to LNL contributes to the robustness performance since it aggregates local information such as lines, edges, shapes, and even objects. In addition, to further improve the robustness performance, we encourage LNL to extract training signal from the moments (a.k.a., mean and standard deviation) and the normalized features. We validate the effectiveness and generality of LNL by achieving state-of-the-art results in terms of accuracy and robustness metrics on German Traffic Sign Recognition Benchmark (GTSRB) and Canadian Institute for Advanced Research (CIFAR-10). More specifically, for traffic sign classification, the proposed LNL yields gains of 1.1\% and ~35\% in terms of clean and robustness accuracy compared to the state-of-the-art studies.
\end{abstract}





\maketitle
\section{Introduction}
Deep Neural Networks (DNNs) are widely deployed in numerous computer vision applications, including object detection~\cite{zhang2020cascaded, zhang2022cctsdb, manzari2022pyramid}, visual tracking~\cite{zhang2022object, zhang2022scstcf}, object recognition~\cite{tourani2020robust}, and action recognition~\cite{dehkordi2021still}, yielding state-of-the-art performance in a broad range of difficult tasks. Due to their widespread success and ability to deploy in sensitive areas, these networks have now become the top choice for deployment in real-world applications, including but not limited to autonomous driving~\cite{asgarian2021fast}, recommender systems~\cite{wang2018enhanced}, health care~\cite{dehkordi2021lightweight}, salient object detection~\cite{mohammadi2020cagnet, noori2020dfnet}, and defense-related applications~\cite{manzari2021robust, tourani2019robust}.

Deep Neural Networks are susceptible to adversarial examples while these generated malicious perturbations are hidden from human vision. Adversarial vulnerabilities have raised concerns about security of computer vision systems, which has led to a variety of studies on robustifying DNNs and defense methods against such attacks. Defense methods try to strengthen the robustness of models in different ways, e.g., carefully designed~\cite{wu2021wider}, stronger data augmentation~\cite{hendrycks2021many, hendrycks2019augmix, yun2019cutmix}, enhanced cost function~\cite{hashemi2022improving}, improved training strategy~\cite{hashemi2021cnn, hashemi2019secure}, and better activation functions or pooling~\cite{zhang2019making}, etc. Although these methods perform well on Convolutional Neural Networks, there is a lack of comparative study to validate that they also keep the effectiveness on vision transformers.

Attention-based transformers have achieved great success in Natural Language Processing (NLP) and computer vision tasks. Vision Transformer (ViT) is the first attention-based image classification model proposed by Dosovitskiy et al.\cite{dosovitskiy2020image}. On a variety of visual tasks \cite{zhu2021visual, jiang2021transgan, touvron2021training}, ViT models have yielded state-of-the-art results by virtue of specific pre-training phase. When trained with considerably large-scale pre-train datasets like JFT-300M, the Vit models can outperform conventional Convolutional Neural Network (CNN) based counterparts. The training is performed by processing the image in patches.

Recently, different studies have demonstrated that vision transformers could gain better robustness than state-of-the-art CNNs with similar computational complexity \cite{benz2021adversarial, bai2021transformers}. However, vanilla vision transformers are vulnerable to adversarial attacks, same as CNN. We aim to robustify ViT models and maintain their state-of-art performance at the same time. To this end, Locality iN Locality was proposed, which is a robust transformer model. To be more specific, we integrate locality to Feed-Forward Network (FFN) of Transformer iN Transformer (TNT) \cite{han2021transformer} by means of depth-wise convolution \cite{li2021localvit,wang2021uformer,zhang2021rest,yu2021glance} instate of multilayer perceptron. Since locality could pertain a wide range of local structures such as edge and shape of image feature, it would contribute to higher robustness performance. In addition, an implicit data augmentation method, called Moment Exchanger (MoEx), is employed to make a better tradeoff between the robustness and accuracy. Finally, the augmented LNL-MoEx model not only improves the robustness accuracy, but also enhances the clean accuracy in comparison with other vision transformer studies.

\begin{figure}[t]
    \clearcaptionsetup{figure}
	\centering
	\includegraphics[width=\linewidth]{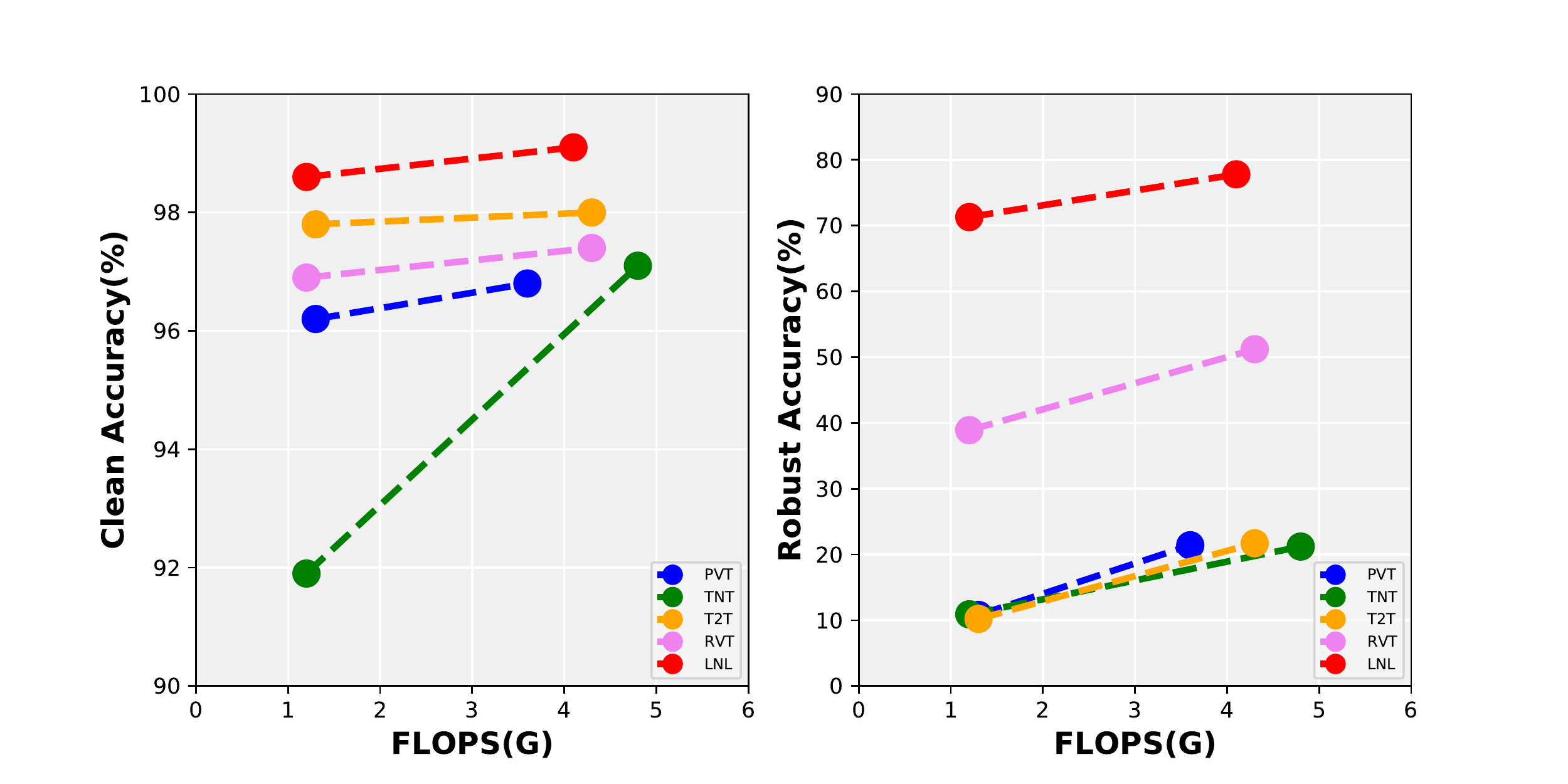}
	\caption{Comparison between LNL and ViT model variants. The robust accuracy is tested on FGSM attack.}
	\label{FIG:1}
\end{figure}

To examine the adversarial robustness of the LNL model, we gauge their performance in terms of the adversarial robustness of vision transformers on traffic sign classification task in white box attack setting shown in Figure~\ref{FIG:1}. We begin with an exhaustive set of experiments to compare the performance of ViT model variants under different perturbations. Various adversarial attack methods have been adopted utilize to generate robust and imperceptible adversarial examples. In this work, two methods have been used, including Fast Gradient Sign Method (FGSM) \cite{goodfellow2014explaining} and Projected Gradient Descent (PGD) \cite{madry2017towards}. The proposed LNL is trained from scratch on the German Traffic Sign Recognition Benchmark (GTSRB) \cite{stallkamp2011german} and Canadian Institute for Advanced Research (CIFAR-10)~\cite{krizhevsky2009learning} without ImageNet~\cite{krizhevsky2012imagenet} pre-training requirement. Our contributions are summarized as follows:

\renewcommand\labelitemi{\small$\bullet$} 
\begin{itemize}
\itemsep=-1pt 		
\itemindent=-3pt 	
\item The depth-wise convolutional filters are integrated in the conventional FFN modules in the transformers (ViT) to account locality principle and gauge its impact on the accuracy/robustness trade-off.
\item To further improve the robustness of our proposed method, an implicit data augmentation method called MoEx is introduce to encourage the model to utilize moment information.
\item The robustness of ViT models are investigated on GTSRB are measured in addition, and the experimental evaluation demonstrates the LNL-MoEx performed better than counterparts.
\end{itemize}

\section{Related Works}
Transformers have made significant contributions to the area of NLP. Thanks to the self-attention module, Transformer can now properly capture the non-local interactions between all different parts of the input sequence, resulting in state-of-the-art performance on a wide range of NLP tasks \cite{brown2020language, dai2019transformer, devlin2018bert, radford2018improving, radford2019language, yang2019xlnet}. The Vision Transformer recently showed that transformers could achieve state-of-the-art performance by pre-training the model on massive data. To this end, the Transformers sequence the input images into patches. ViT requires a computationally expensive pre-training phase on a larger dataset (such as ImageNet-21k \cite{krizhevsky2012imagenet}) because of the lower amounts of inductive biases, as described in \cite{dosovitskiy2020image}, to achieve decent state-of-the-art performance.

Multiple Transformers models have been developed to demonstrate that comparable performance may be achieved without extra data. DeiT \cite{touvron2021training} created a transformer-specific teacher-student technique and trained a transformer architecture only on the ImageNet-1K dataset to relax the requirement of large-scale training dataset in the conventional transformers. Simultaneously, T2T-ViT \cite{yuan2021tokens}, Transformer iN Transformer (TNT) \cite{han2021transformer}, and CvT \cite{wu2021cvt} models have been developed to improve low level feature extractions and further reducing their need on large-scale datasets. These models are known as hybrid-ViTs. Further research is being done to improve the efficiency and performance of transformer architectures by enhancing the ViT architecture \cite{liu2021swin, chu2021conditional}. Our study achieves a high level of performance without a large-scale training by using just GTSRB \cite{stallkamp2011german}.

Concurrent to our work, several recent works analyze the robustness of ViTs from different aspects. Early works focus on the adversarial robustness of ViTs. They demonstrated that ViTs are more robust to adversarial attacks than CNNs. \cite{shao2021adversarial}, and the adversarial transferability between CNNs and ViTs is significantly low \cite{mahmood2021robustness}. Follow-up studies \cite{bhojanapalli2021understanding, paul2021vision, mao2021towards} expand the robustness of ViT models to much common image corruption and distribution shift and prove that ViTs are robust learners. Although several findings are consistent with the above works, in this paper, we do not make a simple comparison of robustness between ViTs and CNNs. We extensively compare the ViT variants architecture from an adversarial robustness standpoint with white-box attack methods. We design a robust vision transformer and introduce LNL-MoEx to further reduce the fragility of ViT models against adversarial attack.

\section{METHODOLOGY}
\begin{figure*}
	\centering
	\includegraphics[width=\textwidth]{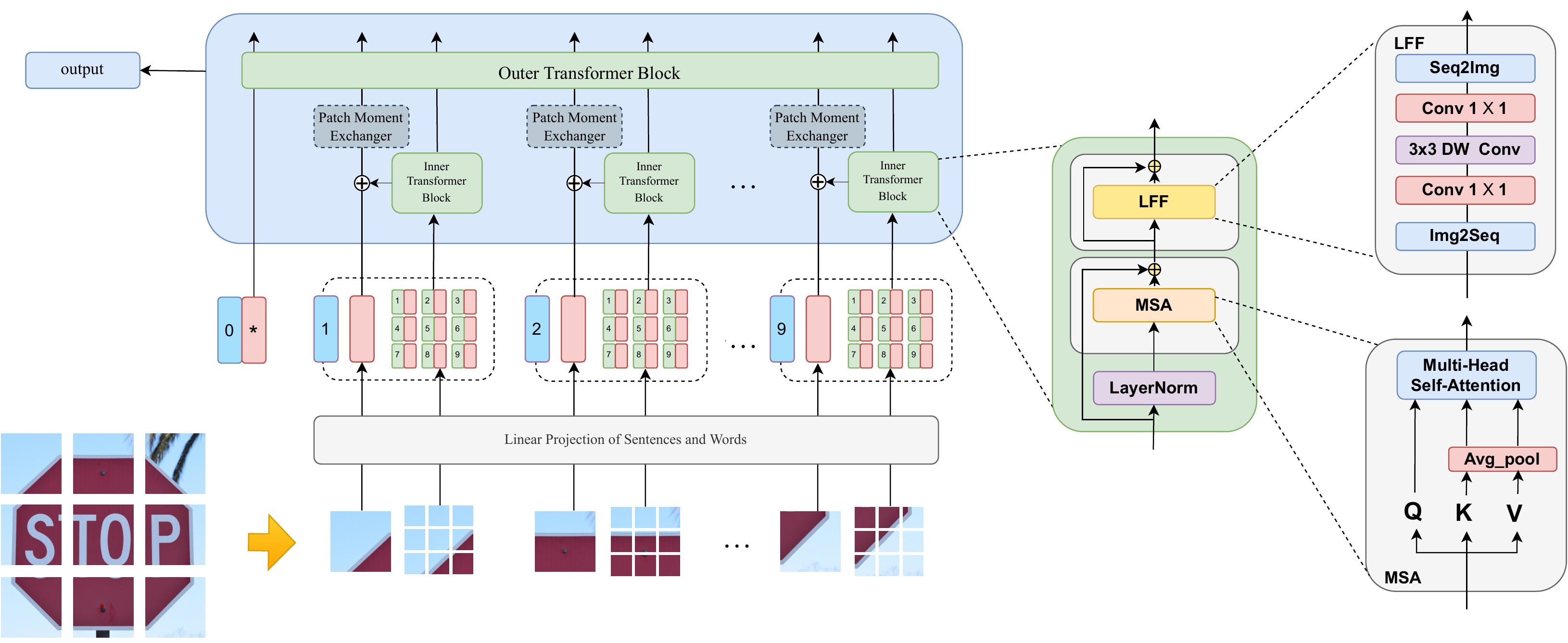}
    \caption{\textbf{Overall architecture of the proposed Locality iN Locality (LNL).} Outer Transformer Block has the same structure as Inner Transformer Block, which is composed of Locally-FeedForward (LFF), Multi-head Self-Attention (MSA), and LayerNorm. Pach Moment Exchanger is our implicit data augmentation method that appears only in the training stage.}
	\label{FIG:2}
\end{figure*}

In this paper, we propose Locality iN Locality (LNL) transformer architecture for visual recognition as illustrated in Figure~\ref{FIG:2}. Firstly, we give a brief overview of the primary components of TNT in subsection~\ref{TNT}. We then explain our proposed Locally FeedForward and build our LNL by adding locality mechanism into the FFN component of TNT. Finally, the Moment Exchanger (MoEx) implicit augmentation is integrated into the proposed LNL model in subsection~\ref{LNL} so that we can encourage the LNL-MoEx to utilize the moment information for enhancing robustness/accuracy tradeoff.

\subsection{Transformer iN Transformer (TNT)}
\label{TNT}

TNT splits a 2D image into $n$ patches $ X = [x^1, x^2,..., x^n]\in \mathbb{R}^{n\times p\times p\times 3} $ uniformly, where $(p, p)$ is the resolution of each image patch. In TNT, the patches are viewed as visual sentences for representing the input images. Then each patch is divided into $m$ sub-patches. That is to say, a visual sentence is composed of a sequence of visual words. This operation can be formulated as follows:

\begin{align}\
  && X^i \rightarrow [x^{i,1}, x^{i,2},..., x^{i,m}],
\end{align}
where $x^{(i,j)}\in \mathbb{R}^{s\times s\times 3}$ is the $j-th$ visual word of the $i-th$ visual sentence, $(s, s)$ is the spatial size of each sub-patches, and $j = 1, 2,..., m$. With a linear projection, the visual words $x^{i,j}$ are transformed into a sequence of word embedding as follows:

\begin{align}\
  && y^{i,j} &= FC(VEC(x^{i,j})),\\
  && Y^i &= [y^{i,1}, y^{i,2},..., y^{i,m}],
\end{align}
where $y^{(i,j)}\in \mathbb{R}^c$ is the $j-th$ word embedding of the $i-th$ visual sentence, $c$ is the dimension of word embedding, and $VEC(.)$ denotes the vectorization function. The TNT method has two data flows across and inside the visual sentences. More specifically, one data flow is employed for visual sentences and the other one is utilized for the visual words inside each sentence. A transformer block is used to investigate the relationship among visual words and their embedding as follows:

\begin{align}\
   && Y_l^{'i} &= Y_{l-1}^i+MSA(LN(Y_{l-1}^i)),\\
   && Y_l^i &= Y_l^{'i}+MLP(LN(Y_l^{'i})),
\end{align}
where the standard transformer blocks consist of a Multi-head Self-Attention module (MSA), a Multiple Layer Perceptron (MLP) and Layer Normalization (LN).  In inner transformer blocks of TNT, $l = (1, 2,..., L)$ indicate the index of the $l-th$ transformer block, and $L$ indicates the overall quantity of stacked blocks. After transformation, the word embedding would be converted as $Y_l=[Y_l^1,Y_l^2,...,Y_l^n]$ , which is transformer block $T_{in}$. This can be viewed as an inner transformer block, denoted as $T_{in}$ process. This operation could account the connections among visual words. The series of sentence are kept in the embedding memory at the sentence level $Z_0=[Z_{class} ,Z_0^1,Z_0^2,...,Z_0^n ]\in \mathbb{R}^{(n+1)\times d}$  In each layer, the word embedding are linear projected onto the sentence embedding domain in each layer as follows:

\begin{align}\label{Eq:6}
   &&Z_{l-1}^i = Z_{l-1}^i+FC(VEC(Y_l^i)),
\end{align}
where $Z_{l-1}^i\in \mathbb{R}^d$ and the FC layer ensures that the dimension is addable. With the above addition operation, Eq. \ref{Eq:6} augments the representation of sentence embedding with word-level features. The sentence embeddings are converted by vanilla transformer block as follows:

\begin{align}
    && Z_l^{'} &= Z_{l-1}+MSA(LN(Z_{l-1} )),\\
    && Z_l &= Z_l^{'}+MLP(LN(Z_l^{'} )). \label{Eq:8}
\end{align}

In short, the visual word embeddings and sentence embeddings be expressed can be as follows:

\begin{align}\
    && Y_l,Z_l=TNT(Y_{l-1},Z_{l-1} ).
\end{align}

In TNT, the inner transformer block represents word relationships for local feature extraction, and the outside transformer block  encodes intrinsic sentence-level context. The TNT is built by stacking the TNT components $L$ times.

\subsection{LNL-MoEx}
\label{LNL}

\textbf{Locally-FeedForward} Previous researches \cite{wu2021cvt, yuan2021incorporating} have demonstrated that the FFN in the classical vision transformer struggles to effectively model local dependencies. However, neighboring pixels are valuable references to provide vital information of local details in input images. To capture local dependency, inspired by previews works \cite{li2021localvit, wang2021uformer}, a depth-wise convolution block is integrated into FFN module in the LNL model. A $(k \times k)$ convolution kernel is utilized in each channel in the depth-wise convolution to properly incorporate locality into the LNL model. For features commutation, the feature inside the $(k \times k)$ kernel is integrated to compute a new feature. To enhance the feature dimension for each token, we first apply a linear projection layer. The tokens are then reshaped into 2D feature maps; as a result, the local information would captured by means of the depth-wise convolution. The features are then flattened to reshape tokens, and the channels are shrunk down via another linear layer to match the channel dimensions of input data. The computation could be represented as:

\begin{align}
  && LOC(Z)=I2S((S2I(Z)\odot W_d )).
\end{align}

\begin{figure}
\vspace{-5pt}
    \centering
    \includegraphics[width=0.5\linewidth]{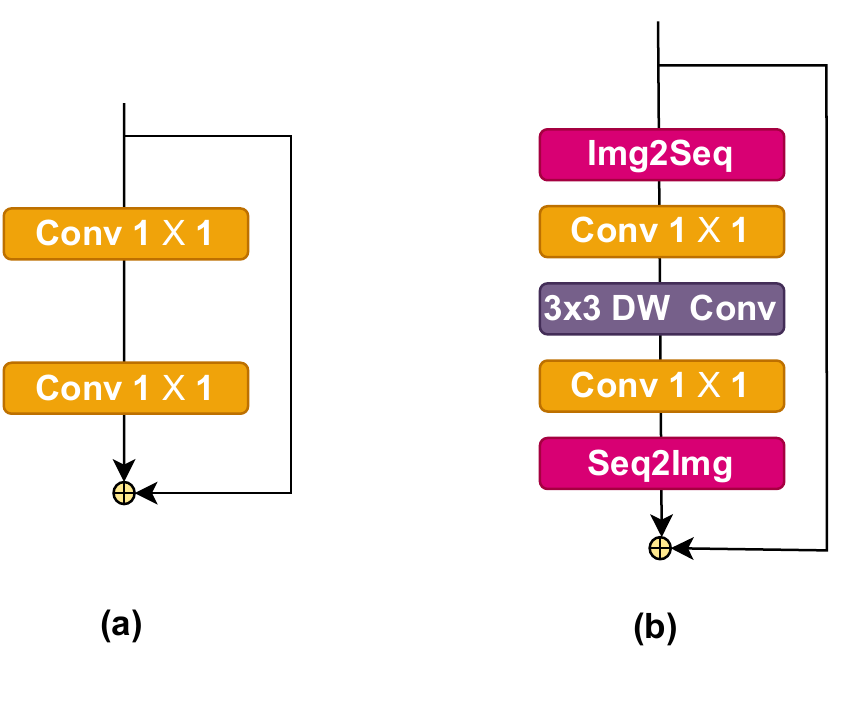}
    \caption{\textbf{Block modifications of Feed Forward Network.} \textbf{(a)} simplified version of Feed Forward Network in TNT. For simplicity of notation, we replace fully-connected layers by 1 × 1 convolutions.  \textbf{(b)} Locally-FeedForward Network, which is composed of depth-wise convolution (DW) and 1 × 1 convolutions to inject locality mechanism into FFN.}
    \label{fig:local}
    \vspace{-1em}
\end{figure}

Where $W_d\in \mathbb{R}^{\gamma d\times 1\times k\times k}$ is the kernel of the depth-wise convolution, $\odot$ is a convolutional operation, $I2S$ is the function that converts an image into a sequence of tokens, and $S2I$ performs the reverse process. Figure~\ref{fig:local} depicts the entire proposed architecture. To introduce locality bias into the TNT, the MLP layer (Eq. \ref{Eq:8}) is replaced by:

\begin{align}
  && Z_l = Z_l^{'} + LOC( Z_l^{'}).
\end{align}

\textbf{MoEx} After feeding the word-level features $Z_l^{'}$ to the depth-wise convolution, the feature representation $Z_l$ would be a 2D tensor. MoEx integrates the feature normalization into the data augmentation, and fuses features with labels across two training samples. The normalized features of one data sample are unified with the feature moments of another. This nonsymmetric combination in feature space aims to encode various directions for the decision boundary. To normalize features in the LNL model, a normalization function $N$ is defined. This function takes the word level features $Z_l^i$ of the $i-th$ input $x_i$ at block $l$ of LNL model. Finally function $N$ generates three outputs which include: the normalized word-level features $\hat{Z}_l^i$, the first-moment $\mu ^i$, and the second moment $\sigma ^i$ as follows:

\begin{align}
  && N(Z_l^i) = ( \hat{Z}_l^i , \mu ^i , \sigma ^i ).
\end{align}

$N$ calculates the value of the first and second momentum after feeding the word level feature $ Z_l^{'} $ through the local module, i.e., depth-wise convolution. This operation relatively resembles PONO function \cite{li2021feature} in the realm of CNNs. Along patch dimension, the normalized features pose zero-mean and standard deviation one. To employ the LNL model, the input images are taken and fused by means of calculated moments. First image $(x_A : Z_A = N^{-1} (\hat{Z}_A , \mu _A , \sigma _A )) $ takes the moments of second image $(x_B : \mu _B , \sigma _B)$  as follows:

\begin{align}
  && Z_A^{(B)} = \sigma _B \frac{Z_A - \mu _A}{\sigma _A} + \mu _B,
\end{align}
where  $ Z_A,\mu _A,\sigma _A $ are the word-level features, the first-moment, and the second moment of image $A$. In addition, $ \mu _B,\sigma _B $ are the first-moment, and the second moment of image $B$. We now proceed with these features $Z_A^{(B)}$. To emphasize on injected
features of $B$, the loss function would be adjusted as follows:

\begin{align}
    &&\lambda \cdot \ell (Z_A^{(B)}, y_A) + (1- \lambda) \cdot \ell (Z_A^{(B)}, y_B),
\end{align}

where $\lambda$ is a fixed variable for setting the combination of the features and the moments $\lambda \in (0,1)$, and $(y_A,y_B)$  are labels of images. Finally, the overall loss function would be the straight-forward combination of the two losses.

\section{EXPERIMENTAL RESULTS}

\subsection{Experimental Settings}

Our experimental evaluations are carried out on the NVIDIA Tesla P100 GPUs. The LNL is assessed under tiny and small model sizes, which are coined LNL-Ti and LNL-S, respectively. Performance of the proposed method is evaluated on German Traffic Sign Recognition Benchmark (GTSRB)~\cite{stallkamp2011german} for focusing on object recognition tasks in autonomous vehicles, and Canadian Institute for Advanced Research (CIFAR-10)~\cite{krizhevsky2009learning} dataset, as a well-known dataset in deep learning. We also implement LNL-MoEx by adding implicit Moment Changer augmentation into the our transformer blocks.\\
The models are all trained on the \textbf{GTSRB} dataset, which contains 39,209 labeled traffic sign images in different 43 classes. The images in the GTSRB dataset are divided into 35,209 training images, 4,000 validation images, and 12,630 test images.\\
\textbf{CIFAR-10} is a subset of the Tiny Images dataset and consists of 60000 (32 x 32) color images. The images are divided into 10 different categories including bird, automobile, airplane, frog, deer, dog, cat, horse, truck, and ship. Train and Test sets of this dataset contain 50000, and 10000 images, respectively. To be more specific, each class in CIFAR-10 provides 5000 training images and 1000 test images. When constructing the dataset, the test set was constructed by selecting 1000 images randomly, and the rest of the images were dedicated to the train set.

All ViT models are trained with the same hyperparameters. The number of training epochs is set to 100 and 150 for GTSRB and CIFAR-10, respectively. The training hyperparameters for the training phase are reported in Table~\ref{table:2}.

\begin{table}[width=.8\linewidth,cols=4,pos=h]
\centering
\begin{threeparttable}
        \caption{Training model parameters for GTSRB and CIFAR-10 datasets.}\label{table:2}
        \vspace{5 pt}
        \begin{tabular*}{\tblwidth}{@{} LLL@{}}
            \toprule
            \textbf{Parameter} & \textbf{GTSRB} & \textbf{CIFAR-10}\\
            \midrule
            Batch Size & 50 &  128\\
            Epoch & 100 & 150\\
            Learning Rate & 0.007 & 0.001\\
            Optimizer & SGD  & SGD\\
            \bottomrule
        \end{tabular*}
\end{threeparttable}
\end{table}

\subsection{LNLs are usable for small dataset }
Table~\ref{table:3} demonstrates the results of the proposed LNL compared to the state-of-the-art studies with respect to the Top-1 and Top-5 robustness accuracy and computational complexity. As reported in Table~\ref{table:3}, the proposed LNL ranks first in terms of clean accuracy and efficiency, and adversarial robustness compared with the state-of-the-art works. To be more specific, the proposed LNL-S yields gains of 0.3\% and 0.5\% on Top-5 metric compared to the second \cite{mao2021towards} and third \cite{wang2021pyramid} best methods. While the obtained gains in comparison with the second-best approach \cite{yuan2021tokens} is not pronounced, our superiority in terms of Top-1 metric is noticeable with a accuracy of 98.2\%. This is due to the fact that the results with respect to the Top-5 metric consider higher probability outputs.

Table~\ref{table:cifar} reports the comparison results on CIFAR-10. The results show that our LNL model significantly outperforms the state-of-the-art methods in terms of clean accuracy. Specifically, we achieve accuracies of 97.6\% and 98.9\% on the validation set using LNL-Ti and LNL-S, respectively. Comparing clean accuracies of GTSRB and CIFAR-10 datasets in Table~\ref{table:3} and Table~\ref{table:cifar} show that traffic sign images are more difficult to predict the correct label. However, it can be seen LNL achieves the best performance on both datasets using fewer or comparable model complexity. These results demonstrate the strong classification ability of the proposed model.

It is worth highlighting that all the counterpart model vision transformers are pre-trained in the ImageNet \cite{krizhevsky2012imagenet} dataset, while our proposed LNL models are only trained on CIFAR-10 and GTSBR datasets from scratch without ImageNet pre-training requirement.

\begin{table*}[width=.9\textwidth,cols=4,pos=t]
\centering
\begin{threeparttable}
	\caption{The performance (\%) of LNL and Transformers on GTSRB and two robustness benchmarks. In Transformers part, models are trained without any extra modifications. In Augmentations part, we trained LNL-Ti model with our proposed MoEx and some augmentation methods such as CutMix.}\label{table:3}
    \vspace{0.2cm}
        \begin{tabular*}{\tblwidth}{@{} C|L|CC|CC|CC@{}}
            \toprule
            \multirow{2}{*}{Group} & \multirow{2}{*}{Model} & \multicolumn{2}{C|}{Model Complexity} & \multicolumn{2}{C|}{GTSRB} & \multicolumn{2}{C}{Robustness Benchmarks}\\
             & & FLOPs (G) & Params (M) & Top-1 & Top-5 & FGSM & PGD\\
            \midrule
            \multirow{11}{*}{Transformers}&PVT-Tiny \cite{wang2021pyramid} & 1.3 & 12.7 & 96.2 & 99.0 & 10.8 & 2.0 \\
            &TNT-T \cite{han2021transformer} & 1.2 & 5.9 & 91.9 & 98.4 & 10.9 & 3.9 \\
            &T2T-ViT-t-10 \cite{yuan2021tokens} & 1.3 & 5.6 & 97.8 & 99.7 & 10.2 & 0.4 \\
            &RVT-Ti \cite{mao2021towards} & 1.2 & 8.6 & 96.1 & 99.1 & 31.5 & 13.9 \\
            &LNL-Ti & 1.2 & 6.1 & \teblebold{97.9} & \teblebold{99.7} & \teblebold{57.7} & \teblebold{37.9} \\
            \cmidrule{2-8}
            &Swin-T \cite{liu2021swin} & 4.1 & 28.5 & 96.8 & 99.1 & 20.2 & 7.8 \\
            &PVT-Small \cite{wang2021pyramid} & 3.6 & 24.0 & 96.8 & 99.3 & 21.4 & 2.4 \\
            &TNT-S \cite{han2021transformer} & 4.8 & 23.4 & 97.1 & 98.9 & 21.2 & 6.8 \\
            &T2T-ViT-t-14 \cite{yuan2021tokens} & 4.3 & 21.1 & 98.0 & 99.2 & 21.7 & 4.9 \\
            &RVT-S \cite{mao2021towards} & 4.3 & 21.9 & 97.2 & 99.5 & 46.1 & 25.3 \\
            &LNL-S & 4.1 & 23.8 & \teblebold{98.2} & \teblebold{99.8} & \teblebold{64.5} & \teblebold{45.7} \\
            \midrule
            \multirow{8}{*}{Augmentations}&DeepAugment \cite{hendrycks2021many} & 1.2 & 6.1 & 97.6 & 99.5 & 65.1 & 42.2 \\
            &CutMix \cite{yun2019cutmix}& 1.2 & 6.1 & 98.5 & 99.5 & 62.6 & 40.1 \\
            &AugMix \cite{hendrycks2019augmix}& 1.2 & 6.1 & 98.3 & 99.4 & 61.7 & 39.8 \\
            &Puzzle-Mix \cite{kim2020puzzle}& 1.2 & 6.1 & 98.5 & \teblebold{99.8} & 67.9 & 44.2 \\
            &RVT-Ti* \cite{mao2021towards} & 1.2 & 10.6 & 96.9 & 99.4 & 38.9 & 20.7 \\
            &LNL-MoEx-Ti & 1.2 & 6.1 & \teblebold{98.6} & 99.7 & \teblebold{71.3} & \teblebold{55.4} \\
            \cmidrule{2-8}
            &RVT-S* \cite{mao2021towards} & 4.3 & 23.0 & 97.7 & 99.6 & 51.2 & 32.9 \\
            &LNL-MoEx-S & 4.1 & 23.8 & \teblebold{99.1} & \teblebold{99.9} & \teblebold{77.8} & \teblebold{59.4}\\
            \bottomrule
        \end{tabular*}
\end{threeparttable}
\end{table*}

\subsection{Adversarial Robustness}
\label{Adversarial}
To evaluate robustness against adversarial attacks, we adopt single-step attack algorithm FGSM \cite{goodfellow2014explaining} and multi-step attack algorithm PGD \cite{madry2017towards} with step number $ t = 5 $, and step size $ \alpha = 0:5 $. The input image is perturbed by both attackers with a maximum magnitude of $1$. Results in Table~\ref{table:3} demonstrate that the adversarial robustness dramatically impacts on the LNL architecture with similar computational complexity, the proposed LNL models represent high adversarial robustness under adversarial attacks. This is ascribed to the proposed modifications on LNLs, which aims to strengthen the adversarial robustness. More specifically, the depth-wise convolution introduces locality inductive Bias into FFN to model local dependencies and introduce inductive bias in favor of better adversarial robustness. Moreover, the MoEx module utilizes implicit data augmentation, which is helpful for adversarial robustness.

Table~\ref{table:3} and Table~\ref{table:cifar} show that the proposed LNL model achieves superior performance on both admired FGSM and PGD attacks in compared state-of-the-art models. In detail, LNL-Ti and LNL-S considerably outperforms the state-of-the-art models with a gain of 32.4\% on FGSM attack and a gain of 34.7\% on PGD attack compared to its counterpart vision transformers on GTSRB dataset. Our model similarly outperforms prior arts by a large margin on both adversarial benchmarks for the CIFAR-10 dataset. Nonetheless, our LNL model generally yields the best accuracy/robustness trade-off.

This advance is further expanded by our MoEx augmentation. Table~\ref{table:3} additionally shows that LNL-MoEx improves both the accuracy and robustness. Compared to other augmentation methods, when LNL-Ti is trained using MoEx achieves the top-1 accuracy of 98.6\% and robust accuracy of 71.3\%, 55.4\% on FGSM and PGD benchmarks. Notably, the highest robust accuracy is obtained by LNL-MoEx-S with 77.8\% on the FGSM. Our enhanced model with MoEx augmentation outperforms other methods by significant improvements on multiple standard benchmarks.

\begin{table}[width=.9\linewidth,cols=4,pos=h]
\centering
\begin{threeparttable}
	\caption{The performance (\%) of LNL and Transformers on CIFAR-10 and two robustness benchmarks.}\label{table:cifar}
     \vspace{-5pt}
        \begin{tabular*}{\tblwidth}{@{} LCRC@{}}
            \toprule
             \multirow{2}{*}{Model} &  \multirow{2}{*}{Top-1. Acc} & \multicolumn{2}{R}{Robustness Benchmarks}\\
             \cmidrule{3-4}
               &  & FGSM & PGD\\
            \midrule
            PVT-Tiny \cite{wang2021pyramid} & 94.2 & 16.6 & 3.1 \\
            TNT-T \cite{han2021transformer} & 94.9 &  13.2 & 4.2 \\
            T2T-ViT-t-10 \cite{yuan2021tokens} & 95.3 & 10.9 & 1.4 \\
            RVT-Ti \cite{mao2021towards} & 95.1 & 34.3 & 14.2 \\
            LNL-Ti  & \teblebold{97.6} & \teblebold{60.8} & \teblebold{38.2} \\
            \midrule
            Swin-T \cite{liu2021swin} & 96.6 & 29.2 & 11.3 \\
            PVT-Small \cite{wang2021pyramid} & 96.8 & 23.1 & 7.3 \\
            TNT-S \cite{han2021transformer} & 98.7 & 29.3 & 11.3 \\
            T2T-ViT-t-14 \cite{yuan2021tokens} & 97.5 & 26.7 & 17.5 \\
            RVT-S \cite{mao2021towards} & 97.4 & 46.3 & 25.9 \\
            LNL-S & \teblebold{98.9} & \teblebold{69.0} & \teblebold{46.9} \\
            \bottomrule
        \end{tabular*}
\end{threeparttable}
\vspace{5pt}
\end{table}

\begin{table*}[width=.85\textwidth,pos=h]
\centering
\begin{threeparttable}
        \caption{Comparison of the attention maps generated by the LNL model and the TNT.(first column) input images; (second and third columns) attention maps generated by models with clean images; and (fourth and fifth columns) attention map generated by models with adversarial images.}\label{table:1}
        \begin{tabular*}{\tblwidth}{@{} C|C|C|C|C@{}}
            \toprule
            \multirow{2}{*}{Input Image} & \multicolumn{2}{C|}{No Attack} & \multicolumn{2}{C}{FGSM Attack}\\
            \cline{2-5}\addlinespace[2pt]
            & TNT \cite{han2021transformer} & LNL & TNT \cite{han2021transformer} & LNL \\
            \midrule
            {\includegraphics[width=1in]{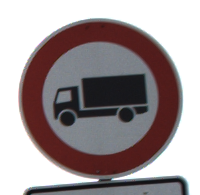}} &
            {\includegraphics[width=1in]{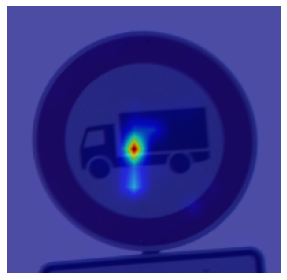}} &
            {\includegraphics[width=1in]{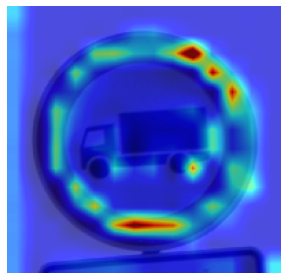}} &
            {\includegraphics[width=1in]{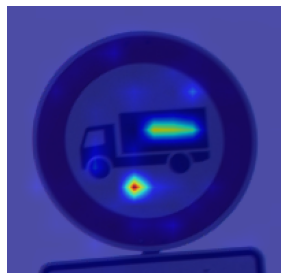}} &
            {\includegraphics[width=1in]{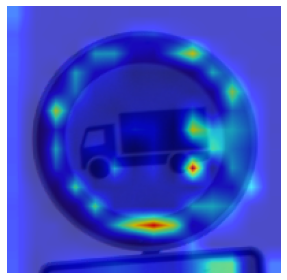}}\\
            \mygray{Label: `over 3.5 tons prohibited'}&
            \mygreen{`over 3.5 tons prohibited' 99\%} &
            \mygreen{`over 3.5 tons prohibited' 99\%} &
            \myred{`No passing' 77\%} &
            \mygreen{`over 3.5 tons prohibited' 79\%}\\
            \midrule
            {\includegraphics[width=1in]{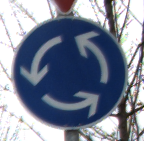}} &
            {\includegraphics[width=1in]{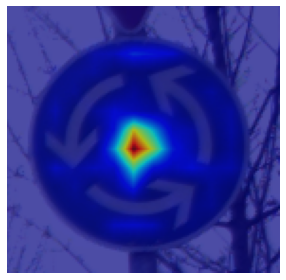}} &
            {\includegraphics[width=1in]{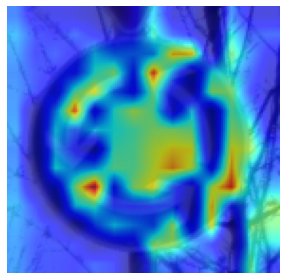}} &
            {\includegraphics[width=1in]{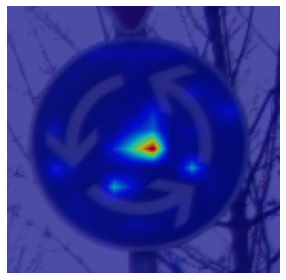}} &
            {\includegraphics[width=1in]{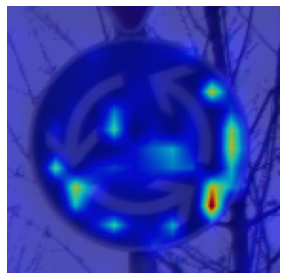}}\\
            \mygray{Label: `Roundabout mandatory'} &
            \mygreen{`Roundabout mandatory' 99\%} &
            \mygreen{`Roundabout mandatory' 99\%} &
            \myred{`Keep right' 99\%} &
            \mygreen{`Roundabout mandatory' 25\%}\\
            \midrule
            {\includegraphics[width=1in]{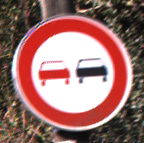}} &
            {\includegraphics[width=1in]{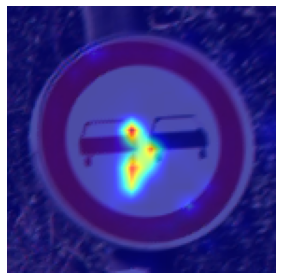}} &
            {\includegraphics[width=1in]{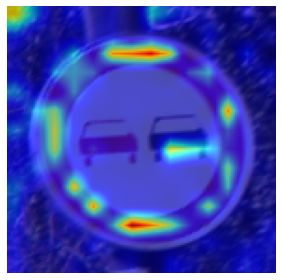}} &
            {\includegraphics[width=1in]{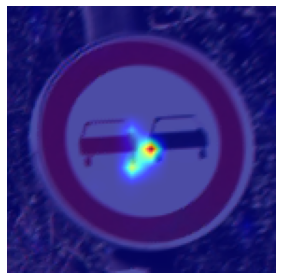}} &
            {\includegraphics[width=1in]{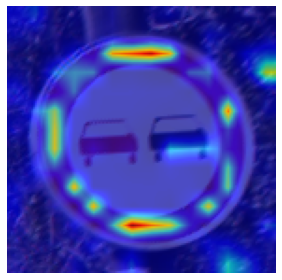}}\\
            \mygray{Label: `No passing'} &
            \mygreen{`No passing' 99\%} &
            \mygreen{`No passing' 99\%} &
            \myred{`Road work' 100\%} &
            \mygreen{`No passing' 98\%}\\
            \midrule
            {\includegraphics[width=1in]{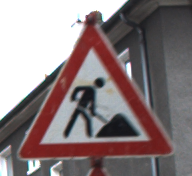}} &
            {\includegraphics[width=1in]{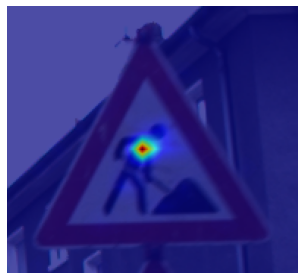}} &
            {\includegraphics[width=1in]{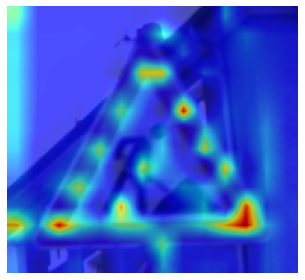}} &
            {\includegraphics[width=1in]{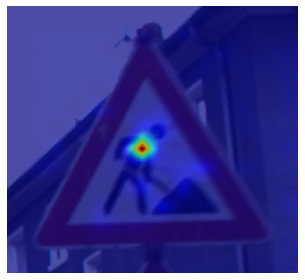}} &
            {\includegraphics[width=1in]{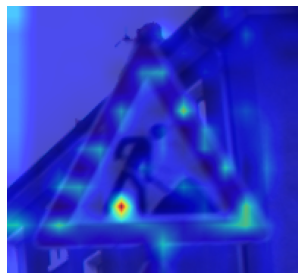}}\\
            \mygray {Label: `Road work'} &
            \mygreen{`Road work' 99\%} &
            \mygreen{`Road work' 99\%} &
            \myred{`Priority road' 99\%} &
            \mygreen{`Road work' 84\%}\\
            \bottomrule
        \end{tabular*}
\end{threeparttable}
\end{table*}

\subsection{Visualizing Attention Maps}
To present qualitative result comparisons to TNT as baseline for our improved model, we compare the attention maps generated by the TNT \cite{han2021transformer} and the LNL using CAM \cite{zhou2016learning}. The attention maps were extracted using a visualization procedure inspired by Caron et al. \cite{caron2021emerging}. In Table~\ref{table:1}, the first column shows the input images of traffic signs from GTSRB \cite{stallkamp2011german}. The second and third columns represent the attention maps of the clean images. In contrast, the fourth and fifth columns represent the attention maps of the adversarial images generated by FGSM \cite{goodfellow2014explaining} under the same settings in subsection~\ref{Adversarial}. Furthermore, the predicted labels are shown with confidence scores for each sample.

From the attention map, we can see TNT just gives attention to widespread features, such as in all cases, it focuses on the center of traffic signs, whereas LNL diverts the attention to more diverse and significant features. For example, in the case of `no passing', `road work' and `Roundabout mandatory' LNL focuses on the frame, color, and shape of traffic signs, respectively, which are important classification features. We can also see from Table~\ref{table:1} that the LNL has strong background separability, and it does not concentrate on trivial background features. It is clear that LNL has the capability to track the local features of the input image while TNT tracks global and centered features.

We can see from the adversarial part that the perturbations highly target the main object of images, the `Roundabout mandatory' for both TNT and LNL. Moreover, the perturbation effect can be noticed on the attention maps. For instance, the FGSM fools the TNT to mislabel `Road work' instate of `No passing' with high confidence. In contrast, the LNL predicts labels with high confidence in almost 3/4 adversarial traffic signs.


\begin{table}[h!]
	\vspace{5pt}
    \small
    \centering
    \caption{\textbf{Effect of Locally FeedForward on other ViT architectures.} Performance (\%) of clean accuracy and adversarial robustness under FGSM attack on GTSRB.}
\label{table:4}
     \tablestyle{5pt}{1.05}
\begin{tabular}{c|cc|l|l}
\toprule

\multirow{2}{*}{Network} & Params& FLOPs & \multirow{2}{*}{Acc} & \multirow{2}{*}{Rob. Acc} \\
& (M) & (G) & & \\
\midrule
T2T-Ti & 5.9 & 1.2 & 97.8 & 10.2 \\
Local-T2T & 5.9 & 1.2 & 99.2 (\textcolor{OrangeRed}{1.4$\uparrow$}) & 33.3 (\textcolor{OrangeRed}{23.1$\uparrow$})\\
\midrule
PVT-Ti & 12.7 & 1.3 & 96.2 & 10.8 \\
Local-PVT & 13.0 & 1.4 & 99.1 (\textcolor{OrangeRed}{2.9$\uparrow$})& 13.0 (\textcolor{OrangeRed}{2.2$\uparrow$})\\
\midrule
Swin-Ti & 28.5 & 4.1 & 96.8 & 20.2 \\
Local-Swin & 28.8 & 4.1 & 98.5 (\textcolor{OrangeRed}{1.7$\uparrow$})& 26.8 (\textcolor{OrangeRed}{6.6$\uparrow$})\\
\bottomrule
\end{tabular}
\vspace{5pt}
\end{table}

\subsection{Ablation Study}
To understand our LNL architecture better, we ablate each critical design by evaluating its performance on GTSRB classification and adversarial benchmarks. Firstly, a study is conducted to demonstrate how the Locally FeedForward could influence the performance of other vision transformers. We also evaluate the performance of our implicit Moment Exchanger augmentation in this section.

\textbf{Impact of Locally FeedForward.} To show the effectiveness of our proposed Locally FeedForward, we introduced local information into the FeedForward network of some transformer architectures. T2T-Ti, PVT-Ti, and Swin-Ti are chosen as the base model. Table~\ref{table:4} illustrates the experimental results of enhanced models by replacing the base FFN module with the proposed Locally FeedForward, all the enhanced models yield significant improvements. Specifically, all the enhanced models achieve more than 2.2\% and 1.4\% promotion on robust and standard accuracy on average. The greatest improvement is for Local-T2T, where the Locally FeedForward leads to a gain of 23\% in robust accuracy.
Compared with the base model, there is only a negligible increase in the amount of computation and a marginal increase in the number of parameters.

\textbf{Impact of Moment Exchanger augmentation.} We further study the impact of MoEx augmentation on multiple transformer blocks. As shown in Table~\ref{table:5}, we can see our augmentation improves standard and robust accuracy of all models. Among them, PVT$^*$ and T2T$^*$  achieve significant improvements of 8.0\% and 1.1\%  on robust and standard accuracy compared to the baselines.

\begin{table}[h!]
	\vspace{5pt}
    \small
    \centering
    \caption{\textbf{Effect of our Patch Moment Exchanger augmentation on other ViT architectures.} Performance (\%) of clean accuracy and adversarial robustness under FGSM attack on GTSRB. Best Results are in bold face.}
\label{table:5}
     \tablestyle{5pt}{1.05}
\begin{tabular}{c|c|c|c|c|c}
\toprule

Vanilla & \multirow{2}{*}{Acc} & \multirow{2}{*}{Rob. Acc} & Improved & \multirow{2}{*}{Acc} & \multirow{2}{*}{Rob. Acc} \\
models &  &  & models & &  \\
\midrule
T2T-Ti & 97.8 & 10.2 & T2T-Ti* & \teblebold{98.1} & \teblebold{16.5} \\
PVT-Ti & 96.2 & 10.8 & PVT-Ti* & \teblebold{96.7} & \teblebold{18.8} \\
Swin-Ti & 96.8 & 20.2 & Swin-Ti* & \teblebold{97.9} & \teblebold{25.7} \\
\bottomrule
\end{tabular}

\vspace{5pt}
\end{table}

\section{CONCLUSION}
We propose the LNL model and study the robustness of vision transformers on traffic sign classification task. The proposed model relaxes the requirement of large-scale pre-training phase in the conventional vision transformer and at the same time outperforms the state-of-the-art transformer-based studies in relation to the clean and adversarial robustness. Furthermore, we integrate the MoEx data augmentation into the vanilla vision transformers to improve adversarial robustness. Instead of disregarding the moments extracted by the normalization layer, the MoEx data augmentation forces the neural network to pay attention towards robust feature. Experimental evaluations approves that the proposed LNL-MoEx consistently achieves outstanding performance on the GTSRB dataset in terms of adversarial robustness and performance clean accuracy. In short, concerning the trade-off between FLOPs, clean and robustness accuracy, extensive experiments validate the superiority of our LNL-MoEx-Ti and LNL-MoEx-S.

\pdfbookmark[section]{CRediT authorship contribution statement}{} 
\printcredits

\nocite{noori2020dfnet, mohammadi2020cagnet, dehkordi2021still,asgarian2021fast, tourani2019robust, tourani2020robust}

\pdfbookmark[section]{Declaration of competing interest / Conflict of interest}{} 
\section*{Declaration of competing interest}
The authors declare that they have no known competing financial
interests or personal relationships that could have appeared
to influence the work reported in this paper.

\pdfbookmark[section]{Acknowledgment}{} 
\section*{Acknowledgment}
we would like to thank the editors and anonymous
reviewers for providing insightful suggestions and comments
to improve the quality of research paper.
\pdfbookmark[section]{References}{} 
\hyphenpenalty=10000 
\bibliographystyle{IEEEtran} 				
\bibliography{cas-refs}

\vskip6pt

\end{document}